\documentclass[sn-mathphys-num,iicol]{sn-jnl}
\usepackage{graphicx}%
\usepackage{caption}%
\usepackage{float}%
\usepackage{subfigure}%
\usepackage{booktabs} 
\usepackage{rotating}   
\usepackage{scalerel}  
\usepackage{multirow}%
\usepackage{amsmath,amssymb,amsfonts}%
\usepackage{amsthm}%
\usepackage{mathrsfs}%
\usepackage[title]{appendix}%
\usepackage{xcolor}%
\usepackage{textcomp}%
\usepackage{manyfoot}%
\usepackage{booktabs}%
\usepackage{lipsum}%
\usepackage{algorithmicx}%
\usepackage{algpseudocode}%
\usepackage{listings}%
\theoremstyle{thmstyleone}%
\theoremstyle{thmstyletwo}%
\theoremstyle{thmstylethree}%
\begin{document}

\title[]{Temporal Attention Evolutional Graph Convolutional Network for Multivariate Time Series Forecasting}
\author[1]{\fnm{Xinlong} \sur{Zhao}}
\author*[1,2]{\fnm{Liying} \sur{Zhang}}\email{lyzhang1980@cup.edu.cn}
\author[1]{\fnm{Tianbo} \sur{Zou}}
\author[1]{\fnm{Yan} \sur{Zhang}}
\affil[1]{College of Information Science and Engineering, China University of Petroleum Beijing 102249, China}
\affil[2]{Beijing Key Laboratory of Petroleum Data Mining, China University of Petroleum Beijing 102249, China}

\abstract{Multivariate time series forecasting enables the prediction of future states by leveraging historical data, thereby facilitating decision-making processes. Each data node in a multivariate time series encompasses a sequence of multiple dimensions. These nodes exhibit interdependent relationships, forming a graph structure. While existing prediction methods often assume a fixed graph structure, many real-world scenarios involve dynamic graph structures. Moreover, interactions among time series observed at different time scales vary significantly. To enhance prediction accuracy by capturing precise temporal and spatial features, this paper introduces the Temporal Attention Evolutional Graph Convolutional Network (TAEGCN). This novel method not only integrates causal temporal convolution and a multi-head self-attention mechanism to learn temporal features of nodes, but also construct the dynamic graph structure based on these temporal features to keep the consistency of the changing in spatial feature with temporal series. TAEGCN adeptly captures temporal causal relationships and  hidden spatial dependencies within the data. Furthermore, TAEGCN incorporates a unified neural network that seamlessly integrates these components to generate final predictions. Experimental results conducted on two public transportation network datasets, METR-LA and PEMS-BAY, demonstrate the superior performance of the proposed model.}

\keywords{Multivariate Time Series Forecasting, Graph Convolutional Network, Multi-head Self Attention, Dynamic Mapping}

\maketitle
\section{Introduction\label{sec1}}

Multivariate Time Series (MTS) encapsulates a continuous time span, reflecting changes in multiple variables over this duration, such as air quality, commodity prices, among others. MTS embodies a collection of data with inherent dependencies, often represented as a graph structure where nodes denote variables.

Classical forecasting methods for MTS encompass statistical models and machine learning techniques. Statistical models, exemplified by the Autoregressive Integrated Moving Average (ARIMA) model, offer high computational efficiency and interpretability across various domains \cite{bib1,bib2}. However, the emergence of deep learning models has garnered significant interest due to their adeptness in nonlinear modeling and resilience to diverse data distributions \cite{bib3}. Consequently, researchers increasingly leverage deep learning approaches for MTS forecasting.

To exploit MTS characteristics fully, some researchers employ Convolutional Neural Networks (CNNs) to capture temporal proximity information, while Recurrent Neural Networks (RNNs) are applied along the temporal axis. For instance, Wu et al. proposed a Convolutional Long Short-Term Fusion Prediction (CLTFP) architecture, combining Long Short-Term Memory (LSTM) and 1-dimensional convolution to predict short-term traffic conditions \cite{bib4,bib5}. Although CLTFP adopts a straightforward approach, it pioneers the alignment of temporal and spatial regularities. However, the rigid structure of regular convolutions confines the model's applicability to grid-based data structures like images or videos, rather than accommodating more diverse structures like graphs. Additionally, RNNs entail iterative training for sequence learning, leading to error accumulation and computational burden.

LSTNet \cite{bib6} and TPA-LSTM \cite{bib7} stand as classical models in MTS prediction, blending convolutional and recurrent neural networks to capture intra- and inter-series correlations. Nonetheless, the non-Euclidean spatial relationships among nodes pose challenges for CNNs' global aggregation in accurately capturing variable correlations.

To address this challenge, recent exploratory research has delved into leveraging Graph Convolutional Networks (GCNs), which are adept at processing non-Euclidean spaces, to tackle issues encountered in multivariate time series (MTS) prediction where Convolutional Neural Networks (CNNs) falter. GCNs have witnessed widespread application in MTS prediction across various domains. They construct a graph structure wherein each MTS constitutes a node, and edges accurately represent interconnections between different nodes, thereby forming a graph structure. Literature \cite{bib8} simplifies the dual-layer program issue and samples discrete graph structures from Bernoulli distributions. Designing suitable graph structures to model correlations between MTS elements and time steps has emerged as a pivotal research focus in this domain. Initially, \cite{bib9} integrated GCNs with gated recursive units to make predictions and introduced a manually crafted adjacency matrix to depict correlations based on node distances. Subsequently, \cite{bib10} contended that predefined graph structures fail to reflect genuine connections and proposed the use of a self-learning adjacency matrix during training. The single-layer GCN serves as a first-order approximation of Chebynet \cite{bib11}, realized by stacking multiple layers to approximate a high-order polynomial filter.

Models for MTS prediction, which comprehend both spatial node characteristics and temporal node evolution, necessitate amalgamating temporal and spatial data information. A model that can concurrently capture temporal and spatial correlations within historical data is termed a Spatio-Temporal Model (STM). Literature \cite{bib10,bib12,bib13} utilized GCNs for spatial information extraction and Temporal Convolutional Network (TCN) for temporal information extraction. Another study \cite{bib14} employed polynomial graph convolution filters and RNNs for extracting temporal dimension information. Additionally, \cite{bib15} proposed a graph attention module for spatial information transfer and combined temporal attention or multi-graph parallel modeling to jointly learn spatio-temporal representations. Spatio-temporal separation models utilize distinct models for extracting temporal and spatial information, whereas spatio-temporal joint models treat time series as directed line graphs that can integrate with graph structures, thereby reducing modeling degrees of freedom \cite{bib16}. Temporal feature extraction in literature \cite{bib17} combines multinomial self-attention with long- and short-term time series analyses.

In scenarios lacking an existing adjacency matrix for a given graph, the graph structure must be constructed initially. However, solely constructing the graph structure based on a particular metric is deemed inadequate \cite{bib18}. Hence, Graph Structure Learning (GSL) assumes a pivotal role in modeling complex networks or graph neural networks. Traditional GSL methods rely on statistical or optimization principles \cite{bib19}, encompassing metric-based methods, probabilistic methods, and direct optimization methods \cite{bib20}. Metric-based methods \cite{bib12,bib14} necessitate initializing an embedding vector for each node, independent of the node's features. During training, the model optimizes this vector and constructs the graph's adjacency matrix using a metric function. While requiring fewer parameters, these methods suffer from slow convergence. On the other hand, the probability-based GSL method \cite{bib21} learns conditional probabilities between node pairs of features, combining them with prior graphs to construct the adjacency matrix, albeit without computational or parameter advantages. Additionally, \cite{bib22} proposed learning graph structures based on temporal sequences. The exploration of using machine learning to jointly infer graph structures and train predictive models in an end-to-end manner is a current focal point.

The complexity inherent in multivariate time series data poses significant challenges to improving prediction accuracy, driving the ongoing development of new prediction methods as a primary research direction. While recent advancements in graph convolutional networks (GCNs) applied to multivariate time series prediction have yielded notable successes through long- and short-term time series analysis and inter-node spatial feature extraction, several critical challenges persist, necessitating further research:

\begin{enumerate}
\item Inefficient Time Dimension Feature Extraction: Efficiently extracting key temporal information from each element within a multivariate time series is crucial. Since individual element sequence data may not adequately characterize the overall dataset, appropriate extraction of temporal information from multivariate time series is imperative. However, multivariate time series often contain redundant features, hindering model extraction efficiency. Hence, designing methods to eliminate redundant information from features is essential to enhance the prediction model's generalization ability.

\item Dynamic Spatial Structure between Nodes: The spatial structure between nodes evolves over time. Existing methods commonly employ a static adjacency matrix throughout, which fails to accommodate temporal changes in node spatial structure. Moreover, existing Graph Structure Learning (GSL) methods are underutilized due to factors like complex computation or slow convergence rates.

\item Variability in Spatial Structure Across Observation Scales: The spatial structure of nodes varies across different observation scales, with correlations differing between short-term and long-term time periods. For instance, in the financial sector, two stocks may exhibit correlated short-term movements due to external factors but diverge in the long term based on internal performance. However, existing methodologies seldom address correlation across different time scales, and fixed graph-structured adjacency matrices cannot adapt to this variability. Consequently, current methods fail to fully exploit the potential of graph neural networks for multivariate time series forecasting problems.
\end{enumerate}

Hence, existing works have yet to fully unleash the potential of graph convolutional networks on forecasting problems. Focusing on traffic information forecasting in a specific area, we design an architecture to integrate the extracted features into a neural network. Our primary contributions are outlined as follows:

\begin{enumerate}
\item This paper leverages the concept of multi-head self-attention and mask mechanism \cite{bib23} to acquire multi-time features, addressing the inefficiency of time information extraction.

\item The paper introduces a novel evolvable graph structure learning method, wherein the graph structure is dynamically updated at each training iteration based on different time periods associated with each node.

\item Inspired by \cite{bib17}, we establish a unified output length for the time features extractor using a Fully-Connected layer after the mask module. This ensures consistency in time length and unifies observation scales across different spatio-temporal layers.
\end{enumerate}

\begin{figure*}
    \centering    \includegraphics[width=0.7\linewidth]{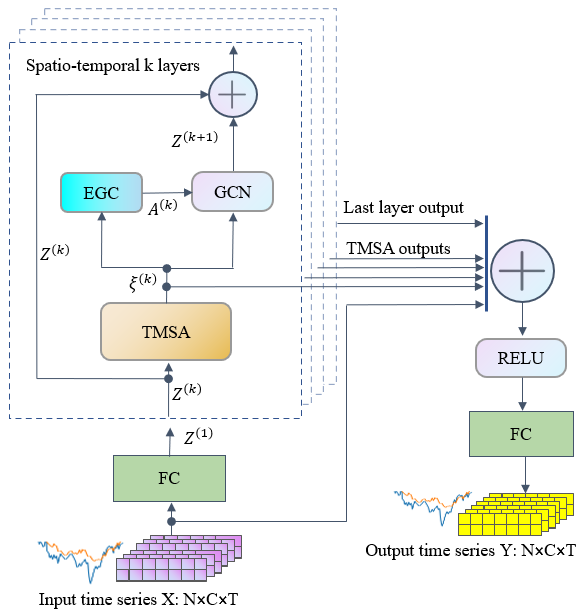}
    \caption{The framework of TAEGCN}
    \label{fig:1}
\end{figure*}

\section{Methodology}\label{sec2}

\subsection{Temporal Attention Evolutional Graph Convolutional Network}\label{subsec1}

We present a novel network architecture for multivariate time series forecasting, termed as the Temporal Attention Evolutional Graph Convolutional Network (TAEGCN). Illustrated in Figure \ref{fig:1}, the model comprises multiple spatio-temporal layers, an output layer, and a fully connected layer. Each spatio-temporal layer integrates Temporal Multi-head Self-Attention (TMSA), Evolvable Graph Construction (EGC), and Graph Convolutional Network (GCN). The overall framework of the model can be defined as follows:

\begin{align}
\xi^{(l)}&=f_{t}^{(l)}(Z^{(l)}) \nonumber  \\
A^{(l)}&=f_{a}^{(l)}(\xi^{(l)})  \\ \label{eq1}
Z^{(l+1)}&=f_{g}^{(l)}(\xi^{(l)},A^{(l)}) \nonumber
\end{align}

In the above formulation, $Z^{(l)}$ represents the input of layer $l$, and a residual neural network is employed to transmit the initialized raw data to the subsequent layer. Here, $f_{t}^{(l)}$ denotes the operation within TMSA, while $f_{a}^{(l)}$ represents the function of the EGC module. The outputs $\xi^{(l)}$ and $A^{(l)}$ correspond to the outputs of the TMSA and EGC modules, respectively. Notably, the parameters of the three modules within the spatio-temporal layer vary across layers, primarily aimed at extracting information of different scales. The output $A^{(l)}$ derived from the EGC module constitutes an adjacency matrix, pivotal for representing the spatial structure within GCN. Lastly, skip connections are incorporated in each layer to transmit data to the final output layer.

TAEGCN innovatively combines features from both temporal and spatial dimensions, effectively extracting features from these two dimensions via TMSA and EGC. Its key advantages are outlined below:

\begin{enumerate}
\item The mask mechanism confines the time series to focus solely on the characteristics within its own block neighborhood. Moreover, the mask window size increases with deeper layers, effectively reducing redundant information and enhancing training efficiency.

\item Within the time dimension, the integration of mask and fully connected layers in each TMSA layer ensures consistency between the output time steps and input. This consistency guarantees uniformity in temporal feature extraction across different layers.

\item Regarding the spatial dimension, the foundation of graph construction lies in the sequential characteristics of various time periods. The EGC module autonomously adjusts node dependencies to derive a more precise graph structure, thereby facilitating accurate multivariate time series prediction.
\end{enumerate}

The subsequent section delves into detailed explanations of the TMSA and EGC modules within the space-time layer.

\begin{figure*}
    \centering
    \includegraphics[width=1\linewidth]{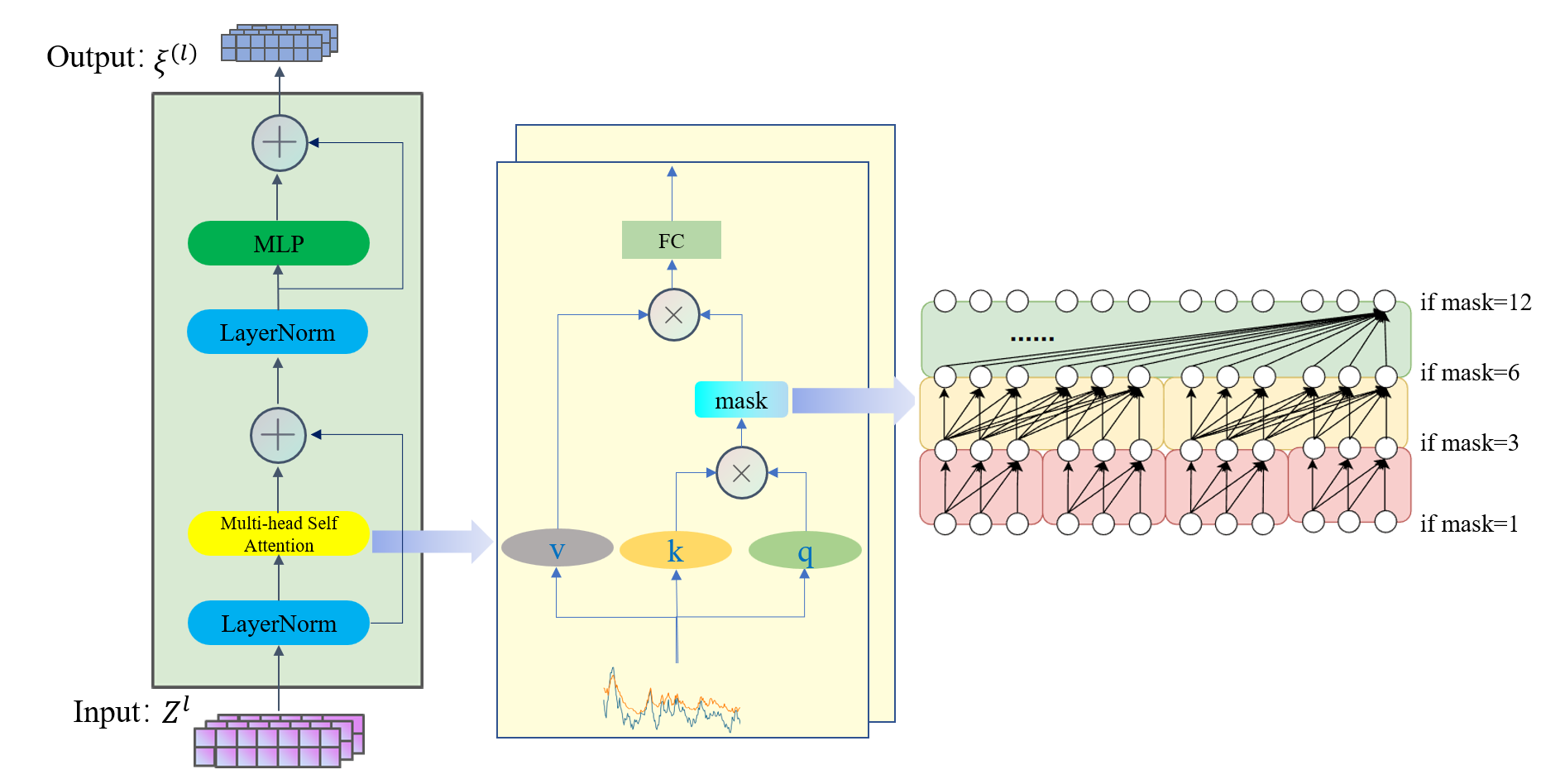}
    \caption{The framework of TMSA, each part adopt 1,3,6,12 windows respectively}
    \label{fig:2}
\end{figure*}

\subsection{Temporal Multi-head Self-Attention}\label{subsec2}
Temporal Multi-head Self-Attention (TMSA) is conceptualized as a dilated temporal convolutional model, drawing inspiration from the multi-head self-attention mechanism \cite{bib21}, depicted in Figure \ref{fig:2}. Within the multi-head self-attention module, data in each head is partitioned into three groups: query ($q$), key ($k$), and value ($v$), which are then processed separately. Subsequently, these processed heads are integrated via matrix cross multiplication and fed into the mask. As the number of layers deepens, the mask's receptive field expands, capturing features across both long and short time steps while ensuring that feature extraction within each period correlates solely with preceding periods—a principle known as the law of temporal causality.

TAEGCN ingeniously leverages TMSA module to effectively extract features from temporal dimension, offering the following advantages:

\begin{enumerate}
\item Consistency in input and output steps ensures multi-step prediction while considering the temporal characteristics of both long and short steps. Given that moments in close proximity typically exhibit stronger correlations than those further apart, the mask mechanism is utilized across different spatio-temporal layers, with distinct window sizes set to differentiate the influence of long and short time steps.

\item TMSA guarantees systematic acquisition of temporal features, ensuring that the current step's sequence value depends solely on data from previous periods, adhering to the principle of temporal causality facilitated by the mask module.
\end{enumerate}

\begin{figure*}
    \centering
    \includegraphics[width=1\linewidth]{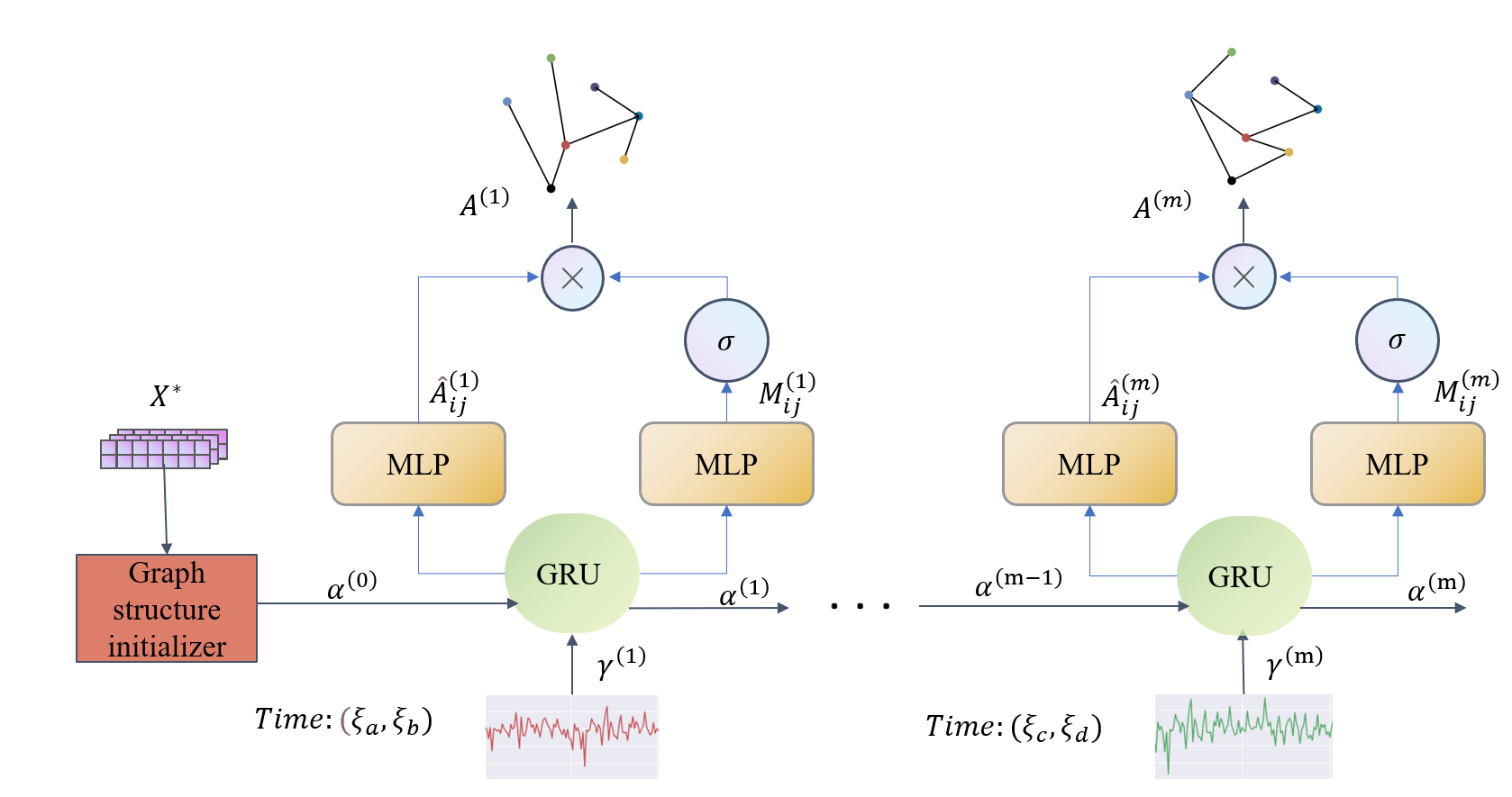}
    \caption{The framework of EGC}
    \label{fig:3}
\end{figure*}

\subsection{Evolvable Graph Construction}\label{subsec3}
The Evolvable Graph Construction (EGC) module serves as an evolutionary graph structure learner, recursively constructing a sequence of adjacency matrices to capture dynamic correlations among variables. EGC not only considers conventional spatial relationships for establishing the graph structure but also incorporates a random stage to explore factors influencing spatial relationship composition, thereby enhancing the capture of hidden spatial dependencies within the data. Based on these features, a more accurate graph structure is established.

The structure of the EGC module, as depicted in Figure \ref{fig:3}, derives the graph structure between nodes in the current period from the previous period's adjacency matrix $A^{(t-1)}$ and the current period's time characteristics, following the relationship:
\begin{equation}
    A^{(t)}=Fe(A^{(t-1)},\xi^{(t)})
\end{equation}
Where $A^{(t)} \in \mathbb{R}^{N \times N}$ represents the adjacency matrix of evolutionary correlation at time $t$, and $\xi^{(t)}$ denotes node features. $Fe$ denotes the function of evolutionary correlation. In practical scenarios, adjacent timestamps typically exhibit temporal consistency, with similar or identical estimates over short durations. Therefore, the model assumes that the graph structure remains constant within a time interval while evolving between adjacent intervals. Additionally, nodes are endowed with an evolving parameter $\alpha$ to mitigate computational costs arising from the $Fe$ function.

The definition of GRU is same as\cite{bib22}, a module for evolving graph representation, is:

\begin{align}
&{r^{(m)}} = \sigma ({W_r}[{\gamma ^{(m)}},{\alpha ^{(m - 1)}}] + {b_r}),\notag\\
&{u^{(m)}} = \sigma ({W_u}[{\gamma ^{(m)}},{\alpha ^{(m - 1)}}] + {b_u}),\label{eq3}\\
&{o^{(m)}} = \mu ({W_o}[{\gamma ^{(m)}},({r^{(m)}} \odot {\alpha ^{(m - 1)}}] + {b_o}),\notag\\
&{\alpha ^{(m)}} = {u^{(m)}} \odot {\alpha ^{(m - 1)}} + (1 - {u^{(m)}}) \odot {o^{(m)}}\notag
\end{align}

Where $r^{(m)}$ and $u^{(m)}$ denote the reset gate and update gate, respectively. $\odot$ represents the element-wise (Hadamard) product, while $W_{r}$, $W_{u}$, and $W_{o}$ denote the learned parameters. $\sigma$ denotes the sigmoid function, and $\mu$ represents the hyperbolic tangent function.

The EGC module integrates these static nodes $\alpha_{s}$ into the fully connected layer, serving as the initial hidden state of the Gated Recurrent Unit (GRU), as shown in the following formula:

\begin{align}
{\alpha ^{(0)}} = ML{P_s}({\alpha _s})
\end{align}

The initialization of the graph structure between nodes is established using data from the multivariate time series itself. This method is adopted due to the inconvenience of acquiring external factors, coupled with the rich information inherent in the multivariate time series dataset. The node feature extractor is employed to extract the static representation ${\alpha _s} \in {R^{N\times{C_s}}}$. Consequently, the initialization graph structure of the global data ${X^{*}}$ obtained through the initializer is as follows:
\begin{align}
{\alpha _{s,i}} = {F_s}(X_i^*)
\end{align}

Here ${\alpha _{s,i}}$ and ${X^{*}}$ represent the static representation and the training data of node $i$, respectively, and ${C_s}$ is the dimension of the static feature. Upon generating the evolved node representations, these two node representations are concatenated, and a multi-layer perceptron is applied to derive the graph structure. Furthermore, a mask is employed to regulate the output message ratio:

\begin{align}
&\hat A_{ij}^{(m)} = ML{P_e}(\alpha _i^{(m)},\alpha _j^{(m)}),\notag\\
&M_{ij}^{(m)} = ML{P_m}(\alpha _i^{(m)},\alpha _j^{(m)}),\\
&{A^{(m)}} = {{\hat A}^{(m)}} \odot \sigma ({M^{(m)}})\notag
\end{align}

${\hat A_{ij}}^{(m)}$ and ${M_{ij}^{(m)}}$ represent the values of the i-th row and j-th column of the graph structure learned by the model, respectively. ${\sigma}$ denotes the sigmoid function, and ${A^{(m)}}$ represents the graph adjacency matrix in the m-th time period, as derived from the final EGC module. Utilizing the graph adjacency matrix and the internal features of each node extracted by TMSA, both datasets are fed into the GCN module to predict the value of the future period. Finally, leveraging the residual network and the fully connected layer, the output result is obtained.

\section{Experiments}\label{sec3}
\subsection{Setup \& Datasets}\label{subsec1}

The parameter settings of the model are as follows: the Adam algorithm is employed as the optimizer for initializing the model parameters, with ${L2}$ regularization applied at a weight of ${10^{-4}}$. The learning rate is set to ${10^{-4}}$, while the batch size is configured to 8, and the training epoch is set to 40 rounds.

\begin{figure}[htbp]
    \centering
    \begin{minipage}[b]{0.5\textwidth}
        \centering
        \includegraphics[width=\linewidth]{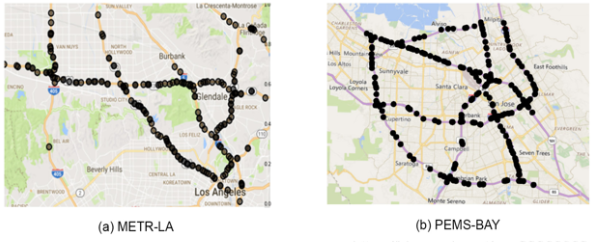}
        \caption{Distribution of METR-LA and PEMS-BAY monitoring sites}
        \label{fig:4}
    \end{minipage}
\end{figure}

For experimental evaluation, this study utilizes two public traffic datasets: METR-LA and PEMS-BAY \cite{bib8}. The configuration of dataset is shown in Table \ref{tab:1}, METR-LA comprises traffic speed and traffic flow statistics from Los Angeles County highways over four months in 2017, while PEMS-BAY encompasses six months of traffic speed and volume data from the San Francisco Bay Area. Figure \ref{fig:4} is the monitor distribution in real map for two datasets. In the data pre-processing phase, sensor readings are aggregated into 5-minute time windows. The dataset is chronologically split, with 70\% allocated for training, 10\% for validation, and 20\% for testing.

The model's parameters are meticulously fine-tuned to optimize performance. The Adam optimizer is employed for parameter initialization, and L2 regularization is applied to mitigate overfitting. The learning rate, batch size, and number of training epochs are determined through a grid search process to ensure convergence towards a stable solution.

\begin{minipage}{.5\textwidth}
    \captionsetup{singlelinecheck=false, justification=justified}
    \captionof{table}{Datasets Summary}
    \scalebox{0.8}{
    \begin{tabular}{ccccc}
    \toprule
    \multicolumn{1}{l}{Dataset} & \multicolumn{1}{l}{Nodes} & \multicolumn{1}{l}{Edges} & \multicolumn{1}{l}{Duration} & \multicolumn{1}{l}{Elements} \\
    \midrule
    METR-LA                     & 207                       & 1515                   & 34272                        & 2 \\
    PEMS-BAY                    & 325                       & 2369                 & 52116                        & 2 \\
    \bottomrule
    \end{tabular}
    }
    \label{tab:1}
\end{minipage}

\subsection{Baselines}
The benchmark model is selected and the following models are used to compare the performance of TAEGCN.
\begin{itemize}
\item ARIMA\cite{bib30}: Auto-Regressive moving average model, which is a traditional time series analysis and forecasting model.
\item FC-LSTM\cite{bib31}: It is a deep learning model that combines a fully connected neural network and a long short-term memory network.
\item WaveNet\cite{bib28} A convolution network architecture for sequence data.
\item DCRNN \cite{bib9} Diffusion convolution recurrent neural network which combines graph convolution networks with recurrent neural networks in an encoder-decoder manner.
\item GGRU\cite{bib29}: Graph Gated Recurrent Unit Network, GGRU uses attention mechanism in graph convolution.
\item STGCN\cite{bib27}: Spatio-Temporal Graph Convolutional Network, which combines graph convolution with 1D convolution.
\item Graph-WaveNet\cite{bib10}: Graph filtering network, combining graph convolutional network and WaveNet's multivariate time series forecasting model.
\end{itemize}

\begin{table*}[ht]
\caption{Performance compasion}
\label{table_2}
\scalebox{0.95}{
\begin{tabular}{@{}cccllllllll@{}}
\toprule
\multirow{2}{*}{Datasets} & \multirow{2}{*}{Models} & \multicolumn{3}{c}{15min}                                                                                    & \multicolumn{3}{c}{30min}                                            & \multicolumn{3}{c}{60min}                       \\ \cmidrule(l){3-11} 
                          &                         & \multicolumn{1}{l}{MAE}           & RMSE                              & \multicolumn{1}{l|}{MAPE}            & MAE           & RMSE          & \multicolumn{1}{l|}{MAPE}            & MAE           & RMSE          & MAPE            \\ \midrule
\multirow{6}*{\rotatebox{90}{METR-LA}}  & ARIMA                   & 3.99                              & \multicolumn{1}{c}{8.21}          & \multicolumn{1}{c|}{9.60\%}          & 5.15          & 10.45         & \multicolumn{1}{l|}{12.70\%}         & 6.90          & 13.23         & 17.40\%         \\
                          & FC-LSTM                 & 3.44                              & \multicolumn{1}{c}{6.30}          & \multicolumn{1}{c|}{9.60\%}          & 3.77          & 7.23          & \multicolumn{1}{l|}{10.90\%}         & 4.37          & 8.69          & 13.20\%         \\
                          & WaveNet                 & 2.99                              & \multicolumn{1}{c}{5.89}          & \multicolumn{1}{c|}{8.04\%}          & 3.59          & 7.28          & \multicolumn{1}{l|}{10.25\%}         & 4.45          & 8.93          & 13.62\%         \\
                          & DCRNN                   & 2.77                              & \multicolumn{1}{c}{5.38}          & \multicolumn{1}{c|}{7.30\%}          & 3.15          & 6.45          & \multicolumn{1}{l|}{8.80\%}         & 3.60          & 7.60          & 10.50\%         \\
                          & GGRU                    & 2.71                              & \multicolumn{1}{c}{5.24}          & \multicolumn{1}{c|}{6.99\%}          & 3.12          & 6.36          & \multicolumn{1}{l|}{8.56\%}          & 3.64          & 7.65          & 10.62\%         \\
                          & STGCN                   & 2.88                              & \multicolumn{1}{c}{5.74}          & \multicolumn{1}{c|}{7.62\%}          & 3.47          & 7.24          & \multicolumn{1}{l|}{9.57\%}          & 4.59          & 9.40          & 12.70\%         \\
                          & Graph-WaveNet           & 2.69                              & \multicolumn{1}{c}{5.15}          & \multicolumn{1}{c|}{6.90\%}          & 3.07          & 6.22          & \multicolumn{1}{l|}{8.37\%}          & 3.53          & 7.37          & 10.01\%         \\
                          & TAEGCN                  & \textbf{2.64}                     & \multicolumn{1}{c}{\textbf{5.03}} & \multicolumn{1}{c|}{\textbf{6.72\%}} & \textbf{2.83} & \textbf{5.33} & \multicolumn{1}{l|}{\textbf{7.70\%}} & \textbf{3.19} & \textbf{6.41} & \textbf{8.73\%} \\ \midrule
\multirow{5}*{\rotatebox{90}{PEMS-BAY}} & ARIMA                   & \multicolumn{1}{l}{1.62}          & 3.30                              & \multicolumn{1}{l|}{3.50\%}          & 2.33          & 4.76          & \multicolumn{1}{l|}{5.40\%}          & 3.38          & 6.50          & 8.30\%          \\
                          & FC-LSTM                 & \multicolumn{1}{l}{2.05}          & 4.19                              & \multicolumn{1}{l|}{4.80\%}          & 2.20          & 4.55          & \multicolumn{1}{l|}{5.20\%}          & 2.37          & 4.96          & 5.70\%          \\
                          & WaveNet                 & 1.39                              & \multicolumn{1}{c}{3.01}          & \multicolumn{1}{c|}{2.91\%}          & 1.83          & 4.21          & \multicolumn{1}{l|}{4.16\%}         & 2.35          & 5.43          & 5.87\%         \\
                          & DCRNN                   & 1.38                              & \multicolumn{1}{c}{2.95}          & \multicolumn{1}{c|}{2.90\%}          & 1.74          & 3.97          & \multicolumn{1}{l|}{3.90\%}         & 2.07          & 4.74          & 4.90\%         \\
                          & STGCN                   & \multicolumn{1}{l}{1.36}          & 2.96                              & \multicolumn{1}{l|}{2.90\%}          & 1.81          & 4.27          & \multicolumn{1}{l|}{4.17\%}          & 2.49          & 5.69          & 5.79\%          \\
                          & Graph-WaveNet           & \multicolumn{1}{l}{1.30}          & 2.74                              & \multicolumn{1}{l|}{2.73\%}          & \textbf{1.63} & 3.70          & \multicolumn{1}{l|}{\textbf{3.67\%}} & \textbf{1.95} & 4.52          & \textbf{4.63\%} \\
                          & TAEGCN                  & \multicolumn{1}{l}{\textbf{1.23}} & \textbf{2.46}                     & \multicolumn{1}{l|}{\textbf{2.48\%}} & \textbf{1.63} & \textbf{3.43} & \multicolumn{1}{l|}{3.73\%}          & 1.98          & \textbf{4.43} & \textbf{4.63\%} \\ \bottomrule
\end{tabular} 
}
\label{tab:2}
\end{table*}

\subsection{Results}
Table \ref{tab:2} presents a comparison of TAEGCN's performance with baseline models on the METR-LA and PEMS-BAY datasets, with unit time lengths of 15 minutes, 30 minutes, and 60 minutes. TAEGCN demonstrates remarkable performance across both datasets, significantly surpassing temporal models like ARIMA and FC-LSTM. Notably, it outperforms previous convolution-based methods such as Graph-WaveNet and recursive-based methods like GGRU. In particular, TAEGCN exhibits performance gains over Graph-WaveNet, the second-best model, across both datasets in the 15-minute to 30-minute horizon. However, performance differences become more pronounced in the 60-minute horizon. While TAEGCN improves performance on the METR-LA dataset, its performance on PEMS-BAY is comparable to that of Graph-WaveNet. Additionally, Table \ref{tab:2} illustrates that as the prediction length of the time series increases, performance declines for both datasets, with the degradation more significant for PEMS-BAY compared to METR-LA.

\begin{figure}[ht]
\centering
\subfigure{\includegraphics[scale=0.24]{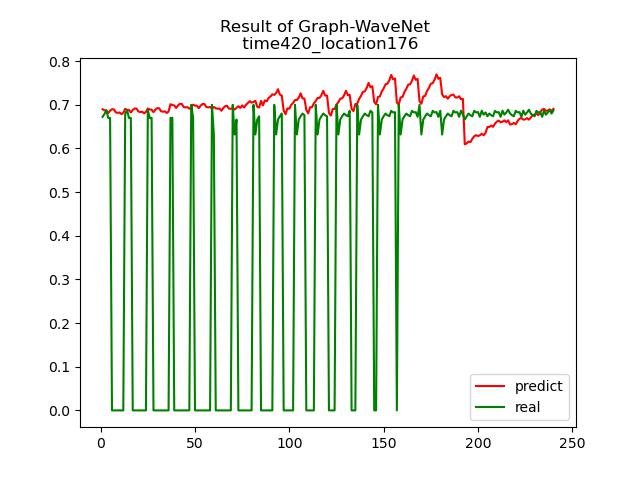}}{\includegraphics[scale=0.24]{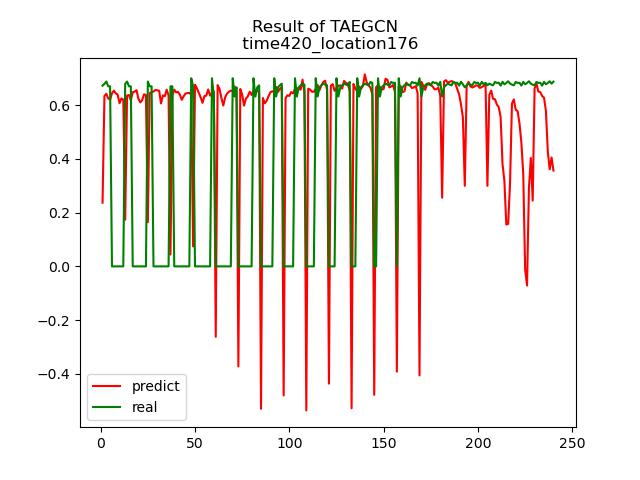}}
\caption{Forecasting Result from Graph-WaveNet and TAEGCN}
\label{fig:5}
\end{figure}

The test results of Graph-WaveNet and TAEGCN models are compared using node 176 and hours 401-420. It is evident that TAEGCN's predicted results exhibit a higher degree of coincidence with the real values compared to Graph-WaveNet. In the left figure of Figure \ref{fig:5}, the predicted values fail to capture the data's volatility, displaying an inconsistent trend with the real values. However, the right figure in Figure \ref{fig:5} depicts simultaneous rises and falls between the predicted and real values during periods of fluctuation, accurately predicting the trend. This indicates that TAEGCN outperforms Graph-WaveNet in multivariate time series forecasting. Subsequently, the paper delves into further exploration of how TAEGCN's temporal feature extractor TMSA and spatial feature extractor EGC modules contribute to performance improvement.

\begin{table*}[ht]
\caption{Ablation result}
\label{table_3}
\scalebox{1.25}{
\begin{tabular}{@{}cccccccc@{}}
\toprule
\multicolumn{1}{l}{\multirow{2}{*}{Dataset}} & \multirow{2}{*}{Models} & \multicolumn{3}{c}{30min}  & \multicolumn{3}{c}{60min} \\ 
\cmidrule(l){3-8} 
\multicolumn{1}{l}{}  &     & MAE  &RMSE  &\multicolumn{1}{c|}{MAPE}  &MAE  &RMSE  &MAPE \\ 
\midrule
\multirow{3}*{\rotatebox{90}{\small {METR-LA}}}                     
& TAEGCN  &\textbf{2.83}  &\textbf{5.33}  &\multicolumn{1}{c|}{\textbf{7.70\%}}  &\textbf{3.19} &\textbf{6.41}  &\textbf{8.73\%} \\
& Ablate TMSA  &2.94  &5.83  &\multicolumn{1}{c|}{7.96\%}  &3.31  & 6.65 &9.28\%  \\
&Ablate EGC    &3.00  &5.71  &\multicolumn{1}{c|}{8.68\%}  &3.45  &7.12  &9.81\%   \\
&Graph-WaveNet &3.07  &6.22  &\multicolumn{1}{c|}{8.37\%}  &3.53  &7.37  &10.01\% \\ 
\midrule
\multirow{3}*{\rotatebox{90}{\small {PEMS-BAY}}}     
& TAEGCN   &\textbf{1.63}  &\textbf{3.43}  &\multicolumn{1}{c|}{3.73\%}  &1.98  &\textbf{4.43} &\textbf{4.63\%} \\
& Ablate TMSA  &1.71  &3.81  &\multicolumn{1}{c|}{3.82\%}  &2.01  &4.47  &4.85\% \\
& Ablate EGC   &1.92  &4.36  &\multicolumn{1}{c|}{4.42\%}  &2.05  &4.60  &4.95\% \\
& Graph-WaveNet  &\textbf{1.63}  &3.70  &\multicolumn{1}{c|}{\textbf{3.67\%}}  &\textbf{1.95}  &4.52  &\textbf{4.63\%}  \\ 
\bottomrule
\end{tabular}
}
\label{tab:3}
\end{table*}

\subsection{Ablation study}

To validate the efficacy of key components, ablation studies were conducted on predictions of the METR-LA and PEMS-BAY datasets at 30 and 60-minute lengths. The temporal feature extractor TMSA and spatial feature extractor EGC modules were individually removed, with the temporal module replaced by a conventional TCN, and the spatial module replaced by a standard GCN. Additionally, the graph structure between nodes was fixed to map points. For comparison, the second-best performing model, Graph-WaveNet, from the benchmark model was selected.

Each experiment utilized identical parameters as TAEGCN, underwent the same number of training cycles, and followed the same data partitioning strategy. Results are presented in Table \ref{tab:3}, yielding the following conclusions:

\begin{enumerate}
\item Both the EGC and TMSA modules contribute to enhancing prediction performance to some degree. Specifically, TAEGCN outperforms Graph-WaveNet which has no these modules, and performs better than models employing either EGC or TMSA alone.

\item Removal of the temporal causal multi-head self-attention module (TMSA) still yields superior performance compared to the benchmark model. This underscores the positive impact of dynamic composition in accurately capturing spatial node relationships across different time periods. The degradation in performance, compared to TAEGCN, highlights its importance in temporal feature extraction. In contrast, the impacts of ablated TMSA are less pronounced than those of ablated EGC in both datasets.

\item Despite competitive results, removal of the Evolvable Graph Structure Learner (EGC) leads to significant performance degradation. This underscores the importance of robust and information-rich causal temporal attention modules in multivariate time series forecasting. The experimental findings underscore the necessity and effectiveness of utilizing the EGC module, as it captures feature information across both short and long time steps, leading to improved predictions. Due to more nodes and more complex connections in PEMS-BAY dataset, the performance of ablation studies in PEMS-BAY is not as good as that in METR-LA.
\end{enumerate}

\subsection{Study of TMSA}

The TMSA module possesses two key characteristics: the local window and temporal causal convolution. The local window feature enables TMSA to allocate more attention weights to adjacent temporal nodes. By stacking time blocks of varying sizes to widen the receptive field and utilizing the self-attention mechanism to fuse neighborhood information, TMSA enhances the coupling of time series information within specific time periods. This coupling facilitates the reflection of time causality. Furthermore, the masking mechanism ensures that the model learns from historical moments within the large receptive field, maintaining the chronological sequence of time. In contrast to traditional TCNs, TMSA ensures the integrity of the time series at each step. This means that the length of the input time series remains consistent with the output, regardless of the size of the dilated convolution kernel or the length of the input. Such characteristics are immensely beneficial in constructing spatial structures based on temporal characteristics. The EGC module receives different time characteristic values, which directly impacts its composition accuracy. Traditional TCNs can only provide time characteristics for single-step predictions, failing to capture the causal relationships within the time series. In contrast, TMSA considers the causality of time series and leverages its broader attention span to extract more accurate temporal features.

\subsection{Study of EGC}
\begin{figure}[ht]
\begin{minipage}[ht]{0.18\textwidth}
\centering
\includegraphics[scale=0.2]{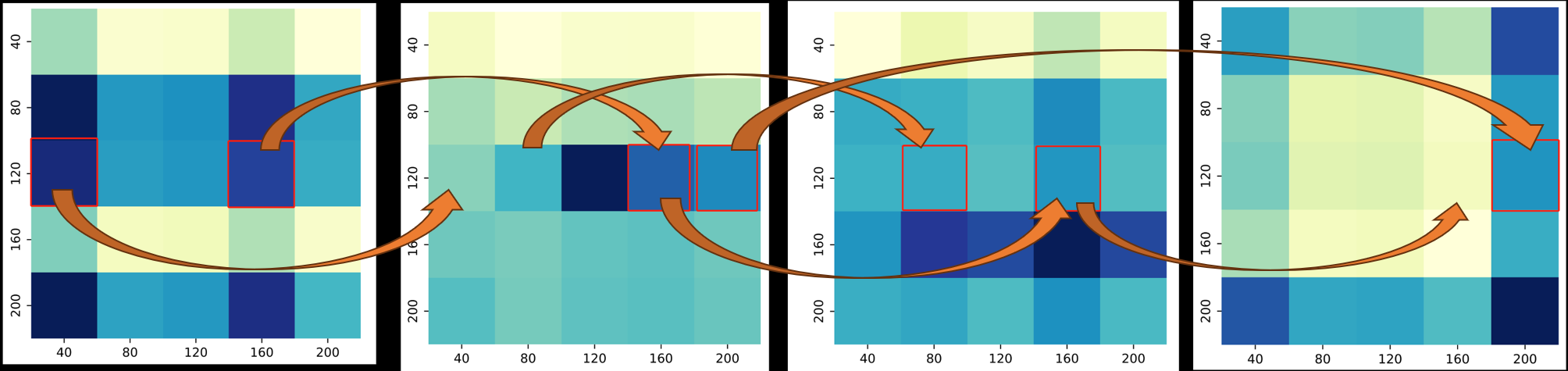}
\end{minipage}
\begin{minipage}[ht]{0.55\textwidth}
\centering
\includegraphics[scale=0.2]{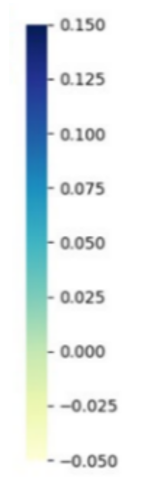}
\end{minipage}
\caption{Spatial Attention in different times and locations}
\label{fig:6}
\end{figure}

To further assess the effectiveness of the EGC module, our team analyzed the spatial dependencies among five monitoring points labeled 40, 80, 120, 160, and 200 in the METR-LA dataset. Figure \ref{fig:6} visualizes the spatial dependence between nodes in the form of a heatmap. In this visualization, blue grids indicate a higher degree of inter-node dependence, while yellow grids represent lower dependence. It's important to note that the adjacency matrix in the heatmap is asymmetrical due to the one-way connections of node dependencies.

\begin{figure}[ht]
\centerline{\includegraphics[scale=0.55]{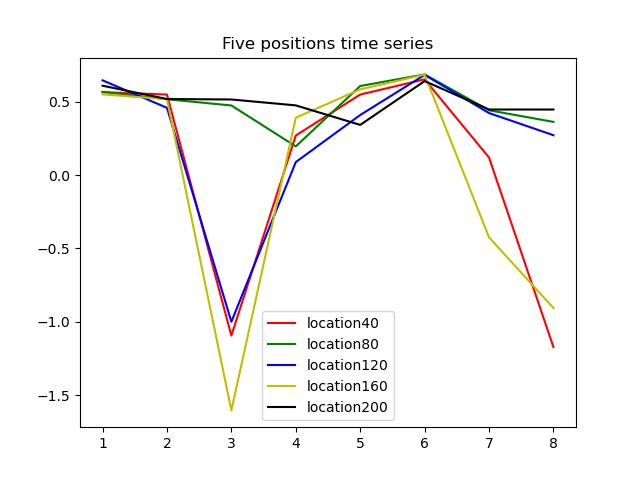}}
\caption{Time series of traffic speed forecasting in different locations}
\label{fig:7}
\end{figure}

Figure \ref{fig:7} presents the original time series curves. Let's consider station 120 as an example and observe some interesting phenomena:

\begin{enumerate}
\item Before time 4, there is a strong correlation between station 120 and stations 40 and 160. The trends of these three stations are similar, as depicted in the first panel of Figure \ref{fig:6}, corresponding to time 3 in the time series curve (Figure \ref{fig:7}). Additionally, the correlation with stations 80 and 200 is notably weaker during this period.

\item The situation changes at times 5 and 6. The trend of station 120 shifts from following stations 40 and 160 to aligning with stations 80 and 200. This transition is clearly evident in the second and third panels of the heatmap, where the colors of stations 80 and 200 transition from light to dark, while the other two stations exhibit the opposite trend.

\item In the fourth panel of Figure \ref{fig:6}, corresponding to time 7, station 120 is only correlated with station 200. At this point, the relationship between station 120 and station 160 has significantly weakened. This change aligns with the pattern observed at time 7 in Figure \ref{fig:7}.
\end{enumerate}

The evolution of correlation from high to low, as depicted by the decreasing values in the adjacency matrix between nodes 120 and 160 over time, aligns well with the trends observed in the time series data (Figure \ref{fig:7}). These findings provide compelling evidence for the efficacy of evolutionary graph structure learners.

\section{Conclusion}
Addressing the limitations of existing multivariate time series forecasting methods in capturing the spatial structure across different time periods and maintaining consistency in the lengths of temporal feature extraction, this study proposes a novel forecasting model: TAEGCN. The model incorporates an evolutionary graph structure learner (EGC) to iteratively construct adjacency matrices that assimilate information from current inputs while retaining historical graph structure information. Additionally, a temporal causal convolutional multi-attention module (TMSA) is introduced to capture time series features across various elements. By amalgamating the outputs of TMSA and EGC modules through a graph convolutional neural network, TAEGCN effectively captures spatio-temporal correlations for improved prediction accuracy. Finally, a unified prediction framework integrates these components to provide the final prediction. Experimental results on real-world datasets demonstrate the superiority of TAEGCN over benchmark models. This study offers a novel model to multivariate time series forecasting, emphasizing the importance of considering spatial structure variations across different time periods and maintaining consistent temporal feature extraction lengths. In future research, we aim to explore graph structure construction methods in diverse scenarios and investigate the applicability of TAEGCN on large-scale datasets. The author expresses gratitude to the anonymous reviewers for their valuable insights and suggestions for enhancing this paper.

\bibliography{sn-bibliography}


\begin{thebibliography}{28}
\ifx \bisbn   \undefined \def \bisbn  #1{ISBN #1}\fi
\ifx \binits  \undefined \def \binits#1{#1}\fi
\ifx \bauthor  \undefined \def \bauthor#1{#1}\fi
\ifx \batitle  \undefined \def \batitle#1{#1}\fi
\ifx \bjtitle  \undefined \def \bjtitle#1{#1}\fi
\ifx \bvolume  \undefined \def \bvolume#1{\textbf{#1}}\fi
\ifx \byear  \undefined \def \byear#1{#1}\fi
\ifx \bissue  \undefined \def \bissue#1{#1}\fi
\ifx \bfpage  \undefined \def \bfpage#1{#1}\fi
\ifx \blpage  \undefined \def \blpage #1{#1}\fi
\ifx \burl  \undefined \def \burl#1{\textsf{#1}}\fi
\ifx \doiurl  \undefined \def \doiurl#1{\url{https://doi.org/#1}}\fi
\ifx \betal  \undefined \def \betal{\textit{et al.}}\fi
\ifx \binstitute  \undefined \def \binstitute#1{#1}\fi
\ifx \binstitutionaled  \undefined \def \binstitutionaled#1{#1}\fi
\ifx \bctitle  \undefined \def \bctitle#1{#1}\fi
\ifx \beditor  \undefined \def \beditor#1{#1}\fi
\ifx \bpublisher  \undefined \def \bpublisher#1{#1}\fi
\ifx \bbtitle  \undefined \def \bbtitle#1{#1}\fi
\ifx \bedition  \undefined \def \bedition#1{#1}\fi
\ifx \bseriesno  \undefined \def \bseriesno#1{#1}\fi
\ifx \blocation  \undefined \def \blocation#1{#1}\fi
\ifx \bsertitle  \undefined \def \bsertitle#1{#1}\fi
\ifx \bsnm \undefined \def \bsnm#1{#1}\fi
\ifx \bsuffix \undefined \def \bsuffix#1{#1}\fi
\ifx \bparticle \undefined \def \bparticle#1{#1}\fi
\ifx \barticle \undefined \def \barticle#1{#1}\fi
\bibcommenthead
\ifx \bconfdate \undefined \def \bconfdate #1{#1}\fi
\ifx \botherref \undefined \def \botherref #1{#1}\fi
\ifx \url \undefined \def \url#1{\textsf{#1}}\fi
\ifx \bchapter \undefined \def \bchapter#1{#1}\fi
\ifx \bbook \undefined \def \bbook#1{#1}\fi
\ifx \bcomment \undefined \def \bcomment#1{#1}\fi
\ifx \oauthor \undefined \def \oauthor#1{#1}\fi
\ifx \citeauthoryear \undefined \def \citeauthoryear#1{#1}\fi
\ifx \endbibitem  \undefined \def \endbibitem {}\fi
\ifx \bconflocation  \undefined \def \bconflocation#1{#1}\fi
\ifx \arxivurl  \undefined \def \arxivurl#1{\textsf{#1}}\fi
\csname PreBibitemsHook\endcsname

\bibitem[\protect\citeauthoryear{Guo et~al.}{2019}]{bib1}
\begin{barticle}
\bauthor{\bsnm{Guo}, \binits{F.}},
\bauthor{\bsnm{Ren}, \binits{L.}},
\bauthor{\bsnm{Jin}, \binits{Y.}},
\bauthor{\bsnm{Ding}, \binits{Y.}}:
\batitle{A dynamic {SVR}--{ARMA} model with improved fruit fly algorithm for the nonlinear fiber stretching process}.
\bjtitle{Natural computing}
\bvolume{18},
\bfpage{747}--\blpage{756}
(\byear{2019})
\end{barticle}
\endbibitem

\bibitem[\protect\citeauthoryear{Liu and Liu}{2009}]{bib2}
\begin{barticle}
\bauthor{\bsnm{Liu}, \binits{Q.}},
\bauthor{\bsnm{Liu}, \binits{J.}}:
\batitle{Research on arima-based multivariate time series neural network forecasting model}.
\bjtitle{Statistics and Decision Making}
\bvolume{11}(\bissue{11}),
\bfpage{23}--\blpage{25}
(\byear{2009})
\end{barticle}
\endbibitem

\bibitem[\protect\citeauthoryear{Lee et~al.}{2021}]{bib3}
\begin{bchapter}
\bauthor{\bsnm{Lee}, \binits{J.-G.}},
\bauthor{\bsnm{Roh}, \binits{Y.}},
\bauthor{\bsnm{Song}, \binits{H.}},
\bauthor{\bsnm{Whang}, \binits{S.E.}}:
\bctitle{Machine learning robustness, fairness, and their convergence}.
In: \bbtitle{Proceedings of the 27th ACM SIGKDD Conference on Knowledge Discovery \& Data Mining},
pp. \bfpage{4046}--\blpage{4047}
(\byear{2021})
\end{bchapter}
\endbibitem

\bibitem[\protect\citeauthoryear{Hochreiter and Schmidhuber}{1997}]{bib4}
\begin{barticle}
\bauthor{\bsnm{Hochreiter}, \binits{S.}},
\bauthor{\bsnm{Schmidhuber}, \binits{J.J.N.C.}}:
\batitle{Long short-term memory}.
\bjtitle{Neural Computation}
\bvolume{9}(\bissue{8}),
\bfpage{1735}--\blpage{1780}
(\byear{1997})
\end{barticle}
\endbibitem

\bibitem[\protect\citeauthoryear{Wu and Tan}{2016}]{bib5}
\begin{botherref}
\oauthor{\bsnm{Wu}, \binits{Y.}},
\oauthor{\bsnm{Tan}, \binits{H.}}:
Short-term traffic flow forecasting with spatial-temporal correlation in a hybrid deep learning framework.
arXiv preprint arXiv:1612.01022
(2016)
\end{botherref}
\endbibitem

\bibitem[\protect\citeauthoryear{Lai et~al.}{2018}]{bib6}
\begin{bchapter}
\bauthor{\bsnm{Lai}, \binits{G.}},
\bauthor{\bsnm{Chang}, \binits{W.-C.}},
\bauthor{\bsnm{Yang}, \binits{Y.}},
\bauthor{\bsnm{Liu}, \binits{H.}}:
\bctitle{Modeling long-and short-term temporal patterns with deep neural networks}.
In: \bbtitle{The 41st International ACM SIGIR Conference on Research \& Development in Information Retrieval},
pp. \bfpage{95}--\blpage{104}
(\byear{2018})
\end{bchapter}
\endbibitem

\bibitem[\protect\citeauthoryear{Shih et~al.}{2019}]{bib7}
\begin{barticle}
\bauthor{\bsnm{Shih}, \binits{S.-Y.}},
\bauthor{\bsnm{Sun}, \binits{F.-K.}},
\bauthor{\bsnm{Lee}, \binits{H.-y.}}:
\batitle{Temporal pattern attention for multivariate time series forecasting}.
\bjtitle{Machine Learning}
\bvolume{108},
\bfpage{1421}--\blpage{1441}
(\byear{2019})
\end{barticle}
\endbibitem

\bibitem[\protect\citeauthoryear{Shang et~al.}{2021}]{bib8}
\begin{botherref}
\oauthor{\bsnm{Shang}, \binits{C.}},
\oauthor{\bsnm{Chen}, \binits{J.}},
\oauthor{\bsnm{Bi}, \binits{J.}}:
Discrete graph structure learning for forecasting multiple time series.
arXiv preprint arXiv:2101.06861
(2021)
\end{botherref}
\endbibitem

\bibitem[\protect\citeauthoryear{Li et~al.}{2017}]{bib9}
\begin{botherref}
\oauthor{\bsnm{Li}, \binits{Y.}},
\oauthor{\bsnm{Yu}, \binits{R.}},
\oauthor{\bsnm{Shahabi}, \binits{C.}},
\oauthor{\bsnm{Liu}, \binits{Y.}}:
Diffusion convolutional recurrent neural network: Data-driven traffic forecasting.
arXiv preprint arXiv:1707.01926
(2017)
\end{botherref}
\endbibitem

\bibitem[\protect\citeauthoryear{Wu et~al.}{2019}]{bib10}
\begin{botherref}
\oauthor{\bsnm{Wu}, \binits{Z.}},
\oauthor{\bsnm{Pan}, \binits{S.}},
\oauthor{\bsnm{Long}, \binits{G.}},
\oauthor{\bsnm{Jiang}, \binits{J.}},
\oauthor{\bsnm{Zhang}, \binits{C.}}:
Graph wavenet for deep spatial-temporal graph modeling.
arXiv preprint arXiv:1906.00121
(2019)
\end{botherref}
\endbibitem

\bibitem[\protect\citeauthoryear{Defferrard et~al.}{2016}]{bib11}
\begin{botherref}
\oauthor{\bsnm{Defferrard}, \binits{M.}},
\oauthor{\bsnm{Bresson}, \binits{X.}},
\oauthor{\bsnm{Vandergheynst}, \binits{P.}}:
Convolutional neural networks on graphs with fast localized spectral filtering.
Advances in neural information processing systems
\textbf{29}
(2016)
\end{botherref}
\endbibitem

\bibitem[\protect\citeauthoryear{Wu et~al.}{2020}]{bib12}
\begin{bchapter}
\bauthor{\bsnm{Wu}, \binits{Z.}},
\bauthor{\bsnm{Pan}, \binits{S.}},
\bauthor{\bsnm{Long}, \binits{G.}},
\bauthor{\bsnm{Jiang}, \binits{J.}},
\bauthor{\bsnm{Chang}, \binits{X.}},
\bauthor{\bsnm{Zhang}, \binits{C.}}:
\bctitle{Connecting the dots: Multivariate time series forecasting with graph neural networks}.
In: \bbtitle{Proceedings of the 26th ACM SIGKDD International Conference on Knowledge Discovery \& Data Mining},
pp. \bfpage{753}--\blpage{763}
(\byear{2020})
\end{bchapter}
\endbibitem

\bibitem[\protect\citeauthoryear{Gu et~al.}{2022}]{bib13}
\begin{barticle}
\bauthor{\bsnm{Gu}, \binits{Z.}},
\bauthor{\bsnm{Chen}, \binits{C.}},
\bauthor{\bsnm{Zheng}, \binits{J.}}, \betal:
\batitle{Traffic flow prediction based on spatio-temporal graph convolutional recurrent neural network}.
\bjtitle{Control and Decision}
\bvolume{37}(\bissue{3}),
\bfpage{645}--\blpage{653}
(\byear{2022})
\end{barticle}
\endbibitem

\bibitem[\protect\citeauthoryear{Bai et~al.}{2020}]{bib14}
\begin{barticle}
\bauthor{\bsnm{Bai}, \binits{L.}},
\bauthor{\bsnm{Yao}, \binits{L.}},
\bauthor{\bsnm{Li}, \binits{C.}},
\bauthor{\bsnm{Wang}, \binits{X.}},
\bauthor{\bsnm{Wang}, \binits{C.}}:
\batitle{Adaptive graph convolutional recurrent network for traffic forecasting}.
\bjtitle{Advances in neural information processing systems}
\bvolume{33},
\bfpage{17804}--\blpage{17815}
(\byear{2020})
\end{barticle}
\endbibitem

\bibitem[\protect\citeauthoryear{Zheng et~al.}{2020}]{bib15}
\begin{bchapter}
\bauthor{\bsnm{Zheng}, \binits{C.}},
\bauthor{\bsnm{Fan}, \binits{X.}},
\bauthor{\bsnm{Wang}, \binits{C.}},
\bauthor{\bsnm{Qi}, \binits{J.}}:
\bctitle{Gman: A graph multi-attention network for traffic prediction}.
In: \bbtitle{Proceedings of the AAAI Conference on Artificial Intelligence},
vol. \bseriesno{34},
pp. \bfpage{1234}--\blpage{1241}
(\byear{2020})
\end{bchapter}
\endbibitem

\bibitem[\protect\citeauthoryear{Pan et~al.}{2020}]{bib16}
\begin{botherref}
\oauthor{\bsnm{Pan}, \binits{C.}},
\oauthor{\bsnm{Chen}, \binits{S.}},
\oauthor{\bsnm{Ortega}, \binits{A.}}:
Spatio-temporal graph scattering transform.
arXiv preprint arXiv:2012.03363
(2020)
\end{botherref}
\endbibitem

\bibitem[\protect\citeauthoryear{Liang et~al.}{2023}]{bib17}
\begin{bchapter}
\bauthor{\bsnm{Liang}, \binits{Y.}},
\bauthor{\bsnm{Xia}, \binits{Y.}},
\bauthor{\bsnm{Ke}, \binits{S.}},
\bauthor{\bsnm{Wang}, \binits{Y.}},
\bauthor{\bsnm{Wen}, \binits{Q.}},
\bauthor{\bsnm{Zhang}, \binits{J.}},
\bauthor{\bsnm{Zheng}, \binits{Y.}},
\bauthor{\bsnm{Zimmermann}, \binits{R.}}:
\bctitle{Airformer: Predicting nationwide air quality in china with transformers}.
In: \bbtitle{Proceedings of the AAAI Conference on Artificial Intelligence},
vol. \bseriesno{37},
pp. \bfpage{14329}--\blpage{14337}
(\byear{2023})
\end{bchapter}
\endbibitem

\bibitem[\protect\citeauthoryear{Guo et~al.}{2020}]{bib18}
\begin{barticle}
\bauthor{\bsnm{Guo}, \binits{K.}},
\bauthor{\bsnm{Hu}, \binits{Y.}},
\bauthor{\bsnm{Qian}, \binits{Z.}},
\bauthor{\bsnm{Sun}, \binits{Y.}},
\bauthor{\bsnm{Gao}, \binits{J.}},
\bauthor{\bsnm{Yin}, \binits{B.}}:
\batitle{Dynamic graph convolution network for traffic forecasting based on latent network of laplace matrix estimation}.
\bjtitle{IEEE Transactions on Intelligent Transportation Systems}
\bvolume{23}(\bissue{2}),
\bfpage{1009}--\blpage{1018}
(\byear{2020})
\end{barticle}
\endbibitem

\bibitem[\protect\citeauthoryear{Roddenberry et~al.}{2021}]{bib19}
\begin{bchapter}
\bauthor{\bsnm{Roddenberry}, \binits{T.M.}},
\bauthor{\bsnm{Navarro}, \binits{M.}},
\bauthor{\bsnm{Segarra}, \binits{S.}}:
\bctitle{Network topology inference with graphon spectral penalties}.
In: \bbtitle{ICASSP 2021-2021 IEEE International Conference on Acoustics, Speech and Signal Processing (ICASSP)},
pp. \bfpage{5390}--\blpage{5394}
(\byear{2021}).
\bcomment{IEEE}
\end{bchapter}
\endbibitem

\bibitem[\protect\citeauthoryear{Zhu et~al.}{2021}]{bib20}
\begin{barticle}
\bauthor{\bsnm{Zhu}, \binits{Y.}},
\bauthor{\bsnm{Xu}, \binits{W.}},
\bauthor{\bsnm{Zhang}, \binits{J.}},
\bauthor{\bsnm{Liu}, \binits{Q.}},
\bauthor{\bsnm{Wu}, \binits{S.}},
\bauthor{\bsnm{Wang}, \binits{L.}}:
\batitle{Deep graph structure learning for robust representations: A survey}.
\bjtitle{arXiv preprint arXiv:2103.03036}
\bvolume{14},
\bfpage{1}--\blpage{1}
(\byear{2021})
\end{barticle}
\endbibitem

\bibitem[\protect\citeauthoryear{Elinas et~al.}{2020}]{bib21}
\begin{barticle}
\bauthor{\bsnm{Elinas}, \binits{P.}},
\bauthor{\bsnm{Bonilla}, \binits{E.V.}},
\bauthor{\bsnm{Tiao}, \binits{L.}}:
\batitle{Variational inference for graph convolutional networks in the absence of graph data and adversarial settings}.
\bjtitle{Advances in neural information processing systems}
\bvolume{33},
\bfpage{18648}--\blpage{18660}
(\byear{2020})
\end{barticle}
\endbibitem

\bibitem[\protect\citeauthoryear{Ye et~al.}{2022}]{bib22}
\begin{bchapter}
\bauthor{\bsnm{Ye}, \binits{J.}},
\bauthor{\bsnm{Liu}, \binits{Z.}},
\bauthor{\bsnm{Du}, \binits{B.}},
\bauthor{\bsnm{Sun}, \binits{L.}},
\bauthor{\bsnm{Li}, \binits{W.}},
\bauthor{\bsnm{Fu}, \binits{Y.}},
\bauthor{\bsnm{Xiong}, \binits{H.}}:
\bctitle{Learning the evolutionary and multi-scale graph structure for multivariate time series forecasting}.
In: \bbtitle{Proceedings of the 28th ACM SIGKDD Conference on Knowledge Discovery and Data Mining},
pp. \bfpage{2296}--\blpage{2306}
(\byear{2022})
\end{bchapter}
\endbibitem

\bibitem[\protect\citeauthoryear{Vaswani et~al.}{2017}]{bib23}
\begin{botherref}
\oauthor{\bsnm{Vaswani}, \binits{A.}},
\oauthor{\bsnm{Shazeer}, \binits{N.}},
\oauthor{\bsnm{Parmar}, \binits{N.}},
\oauthor{\bsnm{Uszkoreit}, \binits{J.}},
\oauthor{\bsnm{Jones}, \binits{L.}},
\oauthor{\bsnm{Gomez}, \binits{A.N.}},
\oauthor{\bsnm{Kaiser}, \binits{{\L}.}},
\oauthor{\bsnm{Polosukhin}, \binits{I.}}:
Attention is all you need.
Advances in neural information processing systems
\textbf{30}
(2017)
\end{botherref}
\endbibitem

\bibitem[\protect\citeauthoryear{Lippi et~al.}{2013}]{bib30}
\begin{barticle}
\bauthor{\bsnm{Lippi}, \binits{M.}},
\bauthor{\bsnm{Bertini}, \binits{M.}},
\bauthor{\bsnm{Frasconi}, \binits{P.}}:
\batitle{Short-term traffic flow forecasting: An experimental comparison of time-series analysis and supervised learning}.
\bjtitle{IEEE Transactions on Intelligent Transportation Systems}
\bvolume{14}(\bissue{2}),
\bfpage{871}--\blpage{882}
(\byear{2013})
\end{barticle}
\endbibitem

\bibitem[\protect\citeauthoryear{Sutskever et~al.}{2014}]{bib31}
\begin{botherref}
\oauthor{\bsnm{Sutskever}, \binits{I.}},
\oauthor{\bsnm{Vinyals}, \binits{O.}},
\oauthor{\bsnm{Le}, \binits{Q.V.}}:
Sequence to sequence learning with neural networks.
Advances in neural information processing systems
\textbf{27}
(2014)
\end{botherref}
\endbibitem

\bibitem[\protect\citeauthoryear{Van Den~Oord et~al.}{2016}]{bib28}
\begin{botherref}
\oauthor{\bsnm{Van Den~Oord}, \binits{A.}},
\oauthor{\bsnm{Dieleman}, \binits{S.}},
\oauthor{\bsnm{Zen}, \binits{H.}},
\oauthor{\bsnm{Simonyan}, \binits{K.}},
\oauthor{\bsnm{Vinyals}, \binits{O.}},
\oauthor{\bsnm{Graves}, \binits{A.}},
\oauthor{\bsnm{Kalchbrenner}, \binits{N.}},
\oauthor{\bsnm{Senior}, \binits{A.}},
\oauthor{\bsnm{Kavukcuoglu}, \binits{K.}}, et al.:
Wavenet: A generative model for raw audio.
arXiv preprint arXiv:1609.03499
\textbf{12}
(2016)
\end{botherref}
\endbibitem

\bibitem[\protect\citeauthoryear{Zhang et~al.}{2018}]{bib29}
\begin{botherref}
\oauthor{\bsnm{Zhang}, \binits{J.}},
\oauthor{\bsnm{Shi}, \binits{X.}},
\oauthor{\bsnm{Xie}, \binits{J.}},
\oauthor{\bsnm{Ma}, \binits{H.}},
\oauthor{\bsnm{King}, \binits{I.}},
\oauthor{\bsnm{Yeung}, \binits{D.-Y.}}:
Gaan: Gated attention networks for learning on large and spatiotemporal graphs.
arXiv preprint arXiv:1803.07294
(2018)
\end{botherref}
\endbibitem

\bibitem[\protect\citeauthoryear{Yu et~al.}{2017}]{bib27}
\begin{botherref}
\oauthor{\bsnm{Yu}, \binits{B.}},
\oauthor{\bsnm{Yin}, \binits{H.}},
\oauthor{\bsnm{Zhu}, \binits{Z.}}:
Spatio-temporal graph convolutional networks: A deep learning framework for traffic forecasting.
arXiv preprint arXiv:1709.04875
(2017)
\end{botherref}
\endbibitem

\end{thebibliography}

\end{document}